\documentclass[journal]{IEEEtran}
\usepackage{amsmath,amsfonts}
\usepackage{algorithmic}
\usepackage{algorithm}
\usepackage{array}
\usepackage[caption=false,font=normalsize,labelfont=sf,textfont=sf]{subfig}
\usepackage{textcomp}
\usepackage{stfloats}
\usepackage{url}
\usepackage{verbatim}
\usepackage{graphicx}
\usepackage{cite}
\begin{document}

\title{Multispectral-Polarization-Infrared Multi-Aperture Camouflaged Vehicle Detection}

\author{Shu Wang$^{1,2}$, Xiong Lin$^{1}$, Yuan Song$^{1}$, Junjie Wang$^{1}$, Feng Huang$^{1}$, Wenxi Liu$^{3}$%
\thanks{$^{1}$School of Mechanical Engineering and Automation, Fuzhou University, Fuzhou 350108, China.}%
\thanks{$^{2}$National Biomedical Imaging Center, College of Future Technology, Peking University, Beijing 100871, China.}%
\thanks{$^{3}$College of Computer and Data Science, Fuzhou University, Fuzhou 350108, China.}%
\thanks{Corresponding authors: Feng Huang and Wenxi Liu. E-mail: huangf@fzu.edu.cn; wenxiliu@fzu.edu.cn.}}

\maketitle

\begin{abstract}
Camouflaged vehicle detection is critically important in modern security and military operations. In contrast to natural camouflage, camouflaged vehicles are intentionally designed to suppress discriminative appearance cues in individual sensing modalities such as visible and infrared imagery, making reliable detection difficult for existing single-modality and simple multimodal methods in complex scenes. Most state-of-the-art multimodal detection approaches rely on strict pixel-level registration and feature-level fusion, which often introduce an inherent trade-off between detection accuracy and system latency. To address these challenges, this paper proposes a spectral–polarization–infrared tri-modal camouflaged vehicle detection framework from the perspective of complementary imaging physics, which is co-designed with a multi-aperture imaging setup. Unlike conventional feature-level fusion methods that depend on precise pixel alignment or registration during data preprocessing, we introduce a consensus-guided decision-level fusion strategy that fuses cross-modal detection results within bounded registration tolerances. This design effectively reduces the reliance on accurate cross-modal registration while maintaining high detection accuracy and real-time performance. To support evaluation, we construct a tri-modal camouflaged vehicle dataset comprising infrared, polarization, and spectral modalities. Experiments on this dataset demonstrate that the proposed method achieves 82.7\% mean Average Precision (mAP) with an end-to-end latency of only 16 ms, outperforming representative feature-level fusion approaches. Furthermore, on the public MID-1K dataset, the proposed framework generalizes to RGB–Depth–Infrared modality combinations without architectural modification, achieving a 5.7\% improvement in mAP@0.5 over the state-of-the-art methods. These results indicate that combining decision-level consensus fusion with multi-aperture multimodal sensing provides a scalable and deployable solution for robust camouflaged vehicle detection.
\end{abstract}

\begin{IEEEkeywords}
Camouflaged target detection, multimodal object detection, spectral imaging, polarization imaging, infrared imaging, decision-level fusion, real-time detection.
\end{IEEEkeywords}

\section{Introduction}
\begin{figure}[!t]
\centering
\includegraphics[width=\linewidth]{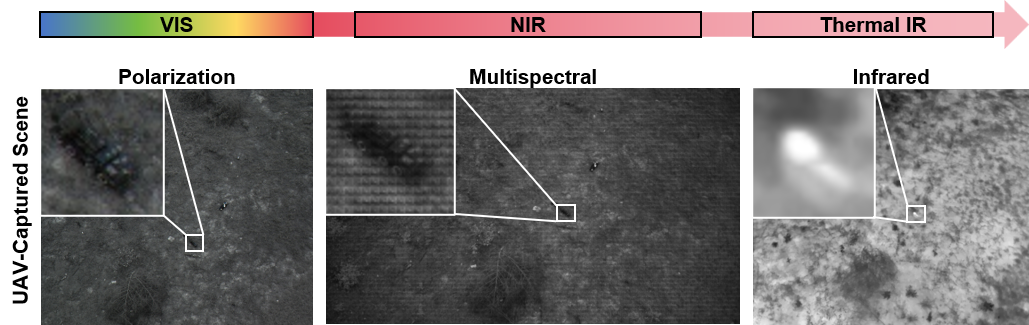}
\caption{UAV-captured unregistered polarization (visible band), multispectral (NIR), and thermal infrared images of a camouflaged vehicle, where the three modalities have different native resolutions before registration and the vehicle appears to be extremely small from aerial view. The top-left corner shows the magnified region of interest where the vehicle locates.}
\label{fig_1}
\end{figure}

\IEEEPARstart{V}{ehicle} detection is a classical task in computer vision. In security and reconnaissance scenarios, vehicles are often intentionally camouflaged, making the reliable localization of camouflaged vehicles critical for threat assessment. Unlike generic Camouflaged Object Detection (COD), which mainly focuses on segmenting visually indistinguishable objects \cite{pang2024zoomnext}, CVD targets vehicles whose visibility is deliberately reduced through coatings, camouflage nets, or thermal regulation, posing distinct challenges in real-world environments. In surveillance and reconnaissance, modern camouflage and concealment strategies like multispectral camouflage coatings, thermal insulation materials, and camouflage nets can suppress discriminative cues in thermal signatures, textures, and structural edges, making vehicles visually and thermally similar to surrounding clutter and challenging both traditional optical systems and learning-based detectors \cite{fan2020camouflaged,zhu2020high}. However, camouflage cannot simultaneously eliminate reflectance, thermal emission, and polarization cues, leading to heterogeneous cross-modal responses to the same camouflaged target (see Fig.~\ref{fig_1}). This problem is further complicated by large intra-class variations across vehicle types (e.g., size, geometry, surface materials, and thermal dynamics), which causes the reliability of each sensing modality to vary across targets and scenes. Moreover, the targets appear small from an aerial perspective (see Fig. 1), which poses an additional challenge.

A practical way to improve visual separability is to exploit vehicle-specific differences in physical properties that are less sensitive to appearance-level camouflage. Since the physical properties of camouflaged vehicles cannot be completely suppressed, the most effective way to access them is through infrared, spectral, and polarization imaging, as these modalities provide the most direct observability of thermal emission, material composition, and surface micro-geometry, respectively. 
By contrast, other sensing modalities mainly capture appearance-level, geometric, or coarse presence cues, such as color and texture in RGB imaging, shape and range in LiDAR, or object-level reflections in radar. These modalities intrinsically provide limited observability of fine-grained physical properties under camouflage. Although jointly acquiring infrared, spectral, and polarization data increases sensing complexity, this cost is intrinsic to camouflage-aware perception, where robust detection depends on complementary physical cues unavailable from any single modality. 

More specifically, multispectral imaging can reveal target-specific reflectance/absorption characteristics of painted coatings, nets, and coverings even when visible contrast is weak \cite{wang2024towards}. Polarization imaging provides complementary cues related to surface roughness and material composition, often enhancing boundary and edge structures of man-made vehicles against natural backgrounds \cite{lyu2023shape}. Infrared imaging captures thermal emission and can highlight vehicles through heat patterns from engines, exhaust regions, or solar-heated surfaces, enabling detection in darkness or under obscurants \cite{liu2024infrared,li2022dense,zhang2022isnet}. These cues are complementary; combining them can improve robustness for multi-type camouflaged vehicles in adverse environments.

Existing multimodal approaches can, in principle, be adapted to tri-modal fusion by combining infrared, spectral, and polarization inputs via early-stage fusion or feature-level integration. Despite encouraging progress in multimodal fusion \cite{li2025pseudo,jeon2024spectral,ying2025visible}, existing pipelines still face two practical limitations for real deployments. First, many methods require pixel-level registration across modalities, which is computationally expensive and often impractical, especially in tri-modal settings, when sensors have different fields of view, resolutions, and viewpoints \cite{liu2024infrared_18,li2025mulfs,sheng2023structure}. Second, directly fusing tri-modal data can be fragile due to severe information imbalance across modalities: high-dimensional spectral inputs may introduce substantial noise and irrelevant variations, which can dominate or dilute discriminative cues when combined with infrared and polarization features \cite{yang2021defending}. 

Inspired by the mantis shrimp’s compound eyes, which employ multiple parallel photoreceptor channels selectively sensitive to different wavelengths and polarization states, we propose a multi-aperture spectral–polarization–infrared imaging and detection framework for real-time multi-type camouflaged vehicle detection. This sensing strategy enables effective discrimination of camouflaged objects in visually complex environments. Specifically, for real-time UAV onboard perception, we develop a UAV-mounted multi-aperture imaging system with multiple relatively independent lenses, which captures synchronized spectral, polarization, and infrared observations within an overlapping field of view and runs on an onboard NVIDIA Jetson AGX Orin platform. We then build modality-specific detectors and, considering that deep multimodal fusion usually incurs high computational cost and inference latency, introduce a consensus-guided decision-level fusion strategy. This strategy first matches the results of two modalities based on bounding-box overlap, then incorporates the third modality for supplementary verification and consistency assessment, and finally produces the target-level decision output according to multimodal consensus. The proposed framework supports real-time multi-target detection in complex environments and remains functional when only a subset of modalities is available. 

The main contributions are summarized as follows:
\begin{itemize}
\item We introduce a new multimodal camouflaged vehicle detection task that exploits spectral, polarization, and infrared observations to recognize camouflaged vehicles in complex environments. We also construct, to the best of our knowledge, the first spectral--polarization--infrared camouflaged vehicle dataset, containing 3,726 synchronized samples over five camouflaged vehicle categories, in which the targets are all small objects occupying only 0.0139\%--0.0305\% of the image area.

\item We propose a snapshot multi-aperture optical system that acquires spectral, polarization, and infrared observations within an overlapping field of view, reducing the need for computationally expensive pixel-level cross-modal registration.

\item  We propose an optics-aware encoding and synchronous detection pipeline, in which the raw observations of each modality are re-encoded to enhance vehicle target-discriminative cues and suppress redundant information, followed by three lightweight unimodal detectors for synchronized real-time detection.

\item We design an efficient plug-and-play decision module that replaces standard non-maximum suppression with cross-modal consensus fusion, associating detections across modalities to determine target locations.

\item Extensive evaluations show that our method gains real-time performance while outperforming the state-of-the-art feature-level fusion methods, achieving 82.7\% mAP@0.5 on multi-target scenarios with 16 ms end-to-end latency.
\end{itemize}

The remainder of this paper is organized as follows. A brief review of related studies is given in Section II. Section III presents the proposed model in detail. Experimental results and corresponding analysis are provided in Section IV. Section V concludes the paper.

\section{Related Work}
This section reviews learning-based camouflage target detection with an emphasis on CVD task, covering (i) spectral, (ii) polarization, (iii) infrared, and (iv) multimodal fusion approaches. We highlight how modality-specific characteristics and practical constraints (e.g., alignment, real-time operation, and modality degradation) motivate our tri-modal spectral–polarization–infrared framework.
\subsection{Spectral Camouflaged Target Detection}
Spectral imaging has been widely explored for camouflage target detection because it can reveal material-dependent spectral responses that remain discriminative even when visible appearance is intentionally matched to the background \cite{hua2015camouflage,yan2019hyperspectral,chen2017camouflage,hupel2022adopting}. Recent deep models primarily improve detection by (1) selecting or emphasizing informative bands and (2) learning spectral–spatial representations tailored to camouflage cues. For example, texture- or boundary-aware designs enhance target–background separability by strengthening boundary details and local contrast \cite{ren2021deep,sun2022boundary}. Band selection/optimization strategies further improve robustness when targets are partially covered (e.g., by camouflage nets) by prioritizing discriminative spectral channels \cite{hupel2023measuring}. Beyond convolutional architectures, representation learning with contrastive objectives and transformer-based modeling has been used to capture long-range spectral–spatial dependencies and explicitly suppress background interference \cite{wu2025contrastive,wang2025hyperspectral}. 

Despite these advances, spectral approaches still face two practical challenges for multi-type camouflaged vehicles. First, spectral measurements are often sensitive to illumination and environmental variations, which can reduce cue reliability in outdoor scenes. Second, the high dimensionality of multispectral data increases computational cost and data bandwidth, making it difficult to simultaneously achieve high accuracy and real-time efficiency in field-deployable systems. These limitations motivate complementary modalities and more efficient multimodal designs.

\subsection{Polarization Camouflaged Target Detection}
Polarization imaging provides additional cues related to surface material, micro-geometry, and reflection properties, which can enhance edge and texture discrimination under challenging visibility conditions \cite{gurton2012remote}. Learning-based polarization processing has evolved along two directions: polarization parameter reconstruction/encoding and task-driven detection/segmentation. Early works integrate convolutional networks with polarization feature extraction to detect artificial targets in outdoor environments \cite{sun2020artificial}. Other studies reconstruct raw polarization measurements to estimate key parameters (e.g., Stokes components and derived quantities such as DoLP and AoP), enabling downstream recognition tasks \cite{zeng2019end}. For low-visibility scenarios, combining polarization-derived cues (e.g., DoLP and S0) with deep networks has shown clear gains in semantic segmentation and scene understanding \cite{wang2022vehicle}. In parallel, physically grounded models demonstrate that polarization encodes rich geometric and material information, and can support robust inference under realistic imaging conditions \cite{lyu2023shape}.

However, polarization cues can be surface- and material-dependent, and their effectiveness may vary across vehicle types and camouflage configurations. In particular, polarization signatures are often stronger for surfaces with pronounced specular reflection or distinct material properties, while common camouflage fabrics/nets may weaken discriminative polarization contrast. These observations suggest that polarization is most effective when combined with complementary modalities (e.g., spectral and infrared) to improve robustness across targets and environments.

\subsection{Infrared Camouflaged Target Detection}
Infrared (IR) imaging is a widely used sensing modality in reconnaissance because it exploits thermal radiation differences between targets and backgrounds, partially bypassing visible-spectrum camouflage such as paint or vegetation cover \cite{zhang2025research}. For camouflaged vehicles, thermal patterns from engines, exhaust regions, tracks, or solar-heated structures can provide discriminative cues, especially in darkness or under obscurants. Consequently, integrating IR imagery with deep models has become a common strategy for improving recognition and localization in complex scenes. 

To strengthen target saliency, many IR detection methods incorporate attention mechanisms, multi-scale feature fusion, and task-driven priors. For instance, edge-aware and bidirectional attention aggregation modules enhance target–background contrast and improve shape reconstruction \cite{zhang2022isnet}. Scale- and location-sensitive losses explicitly emphasize small targets and precise localization, improving detection performance in cluttered infrared scenes \cite{liu2024infrared}. Dense nested U-Net variants with attention-driven multi-scale fusion further preserve weak target signals and enhance robustness for small or low-contrast targets \cite{li2022dense}. Overall, these approaches demonstrate that attention and task priors can substantially improve IR feature representation.

Nevertheless, advanced camouflage materials (e.g., anti-IR coatings or thermal management strategies) can reduce thermal contrast by altering or segmenting target heat signatures, potentially degrading IR-based detection. This motivates the use of additional modalities and robust fusion strategies that remain effective when IR cues are weakened. 

\subsection{Multimodal Fusion for Camouflaged Target Detection}
Single-modality detectors often struggle in complex scenes due to the high similarity between camouflaged targets and backgrounds. Multimodal fusion improves robustness by leveraging complementary cues across modalities. Most existing work focuses on visible–infrared fusion under the RGB–T framework \cite{shen2024icafusion,chen2024deyolo,chen2022multimodal}. Depending on the fusion level, current methods can be broadly grouped into pixel-level, feature-level, and decision-level fusion.

Pixel-level fusion is structurally simple (e.g., channel concatenation) but typically requires strict cross-modal alignment, limiting deployment in non-custom or multi-sensor systems. Feature-level fusion improves performance through dedicated architectures that enable intra- and inter-modal interaction. Representative designs include transformer-based dual-stream aggregation networks for cross-modal feature exchange \cite{qingyun2021cross}, cross-attention fusion with parameter sharing to balance accuracy and efficiency \cite{shen2024icafusion}, and YOLO-style detectors enhanced with spatial/channel attention to mitigate modal interference \cite{chen2024deyolo}. However, many of these methods still assume well-aligned dual-modal inputs and introduce non-trivial computation, which can hinder engineering applicability under real-time constraints.

Decision-level fusion relaxes alignment requirements by combining modality-specific predictions. Probabilistic or Bayesian strategies, for example, can maintain system stability when some modalities are missing or degraded \cite{chen2022multimodal}. Beyond RGB–T, the fusion of polarization and multispectral modalities has also been explored to exploit complementary advantages in material differentiation and spectral response \cite{wu2023object}. In general, multimodal fusion improves accuracy and stability in highly similar backgrounds and complex scenes \cite{zhao2023tmtnet}. Despite this progress, existing camouflage detection systems predominantly consider dual-modality settings and often rely on pre-registered data, leaving practical challenges unresolved for real deployments (e.g., heterogeneous sensors, imperfect alignment, modality dropouts). These gaps motivate our tri-modal spectral–polarization–infrared decision-level fusion framework tailored to efficient and robust detection of multi-type camouflaged vehicle targets in complex scenarios.

\section{Methodology}
\begin{figure}[!t]
\centering
\includegraphics[width=\linewidth]{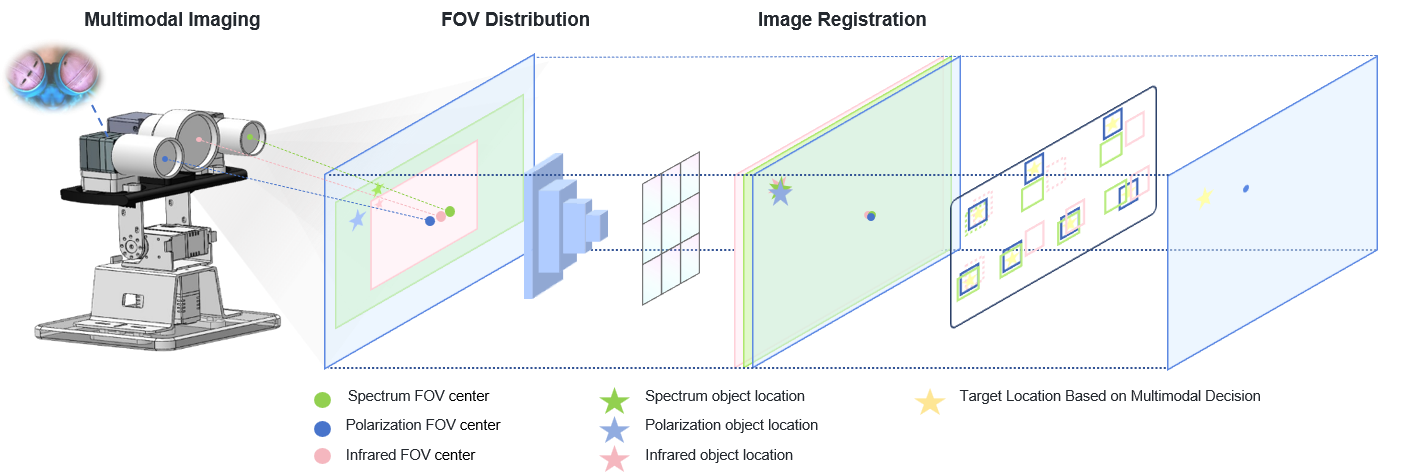}
\caption{Schematic illustration of the multimodal spectral–polarization–infrared camouflaged vehicle recognition and decision-making pipeline. The system first acquires synchronized spectral, polarization, and infrared images with overlapping fields of view using a multi-aperture imaging device; detection and cross-modal image registration are then performed; finally, the detection results from different modalities are combined to achieve target fusion and decision output.}
\label{fig_2}
\end{figure}

Fig.~\ref{fig_2} provides an overview of the proposed multimodal spectral–polarization–infrared camouflaged vehicle detection framework, illustrating the overall pipeline and the relationships among multimodal imaging, detection, and decision-level fusion.

\subsection{Spectral-Polarization-Infrared Multimodal Imaging System}
\begin{figure}[!t]
\centering
\includegraphics[width=\linewidth]{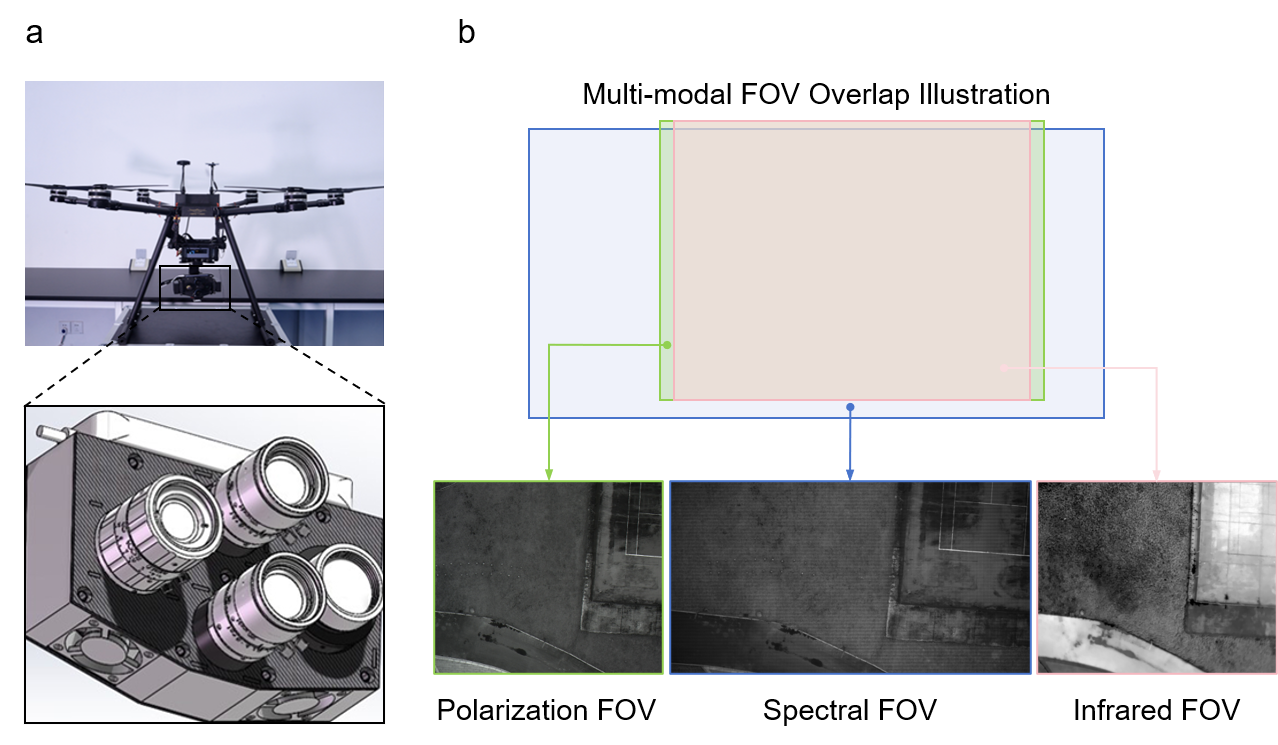}
\caption{UAV-mounted spectral–polarimetric–infrared multimodal imaging system. (a) Photograph of the UAV platform and schematic of the multimodal imaging payload. (b) Illustration of the fields of multimodal FOV overlap as well as their corresponding captured images.}
\label{fig_3}
\end{figure}

As shown in Fig.~\ref{fig_3}, the proposed multimodal imaging system serves as the front-end of the overall framework and is designed to achieve real-time perception of camouflaged targets in UAV scenarios. To this end, we build a UAV-mounted multi-aperture imaging system consisting of four types of sensors, namely multispectral, polarization, infrared, and visible-light cameras. Among them, the multispectral, polarization, and infrared modalities are used for target detection and fusion, while the visible-light modality is retained only as an auxiliary reference.
Let the modality set be defined as
\[
\mathcal{M}=\{sp,po,ir\},
\]
then the multimodal observation at time $t$ can be expressed as
\[
X_t=\{X_t^{sp},X_t^{po},X_t^{ir}\}.
\]

Both the multispectral and polarization cameras operate in snapshot mode, which captures all spectral channels or polarization states within a single exposure rather than scanning them sequentially over time\cite{wunsch2024realisation}. This design reduces acquisition latency, improves data collection efficiency, and mitigates cross-channel misalignment caused by platform or target motion, thereby supporting subsequent real-time target detection.

The multispectral camera operates in snapshot mode and captures reflectance variations across multiple spectral bands, which can expose material-dependent differences between camouflaged vehicles (e.g., coverings and camouflage nets) and the surrounding background. The polarization camera, also in snapshot mode, measures polarization information at multiple analyzer angles (i.e., different polarization states), providing complementary cues related to surface material and edge structures that are often weak in intensity images. The infrared camera records thermal radiation and highlights targets via temperature differences between the vehicle and the background.

Our multi-aperture design enables the cameras to observe an overlapping field of view, which can be coarsely aligned by fine-tuning each camera’s orientation during system setup. To ensure temporal consistency across modalities, we employ a software triggering mechanism to synchronize the frame rates of all image streams, yielding time-aligned multimodal inputs for subsequent fusion and detection. These capabilities support the construction of our multimodal camouflaged-target dataset. The key imaging parameters of the multispectral, polarization, and infrared cameras are summarized in Table~\ref{table_1}.

\begin{table*}[!t]
\renewcommand{\arraystretch}{1.2}
\caption{Camera Parameters of the Multimodal Imaging System}
\label{table_1}
\centering
\begin{tabular}{c c c c c c}
\hline
Camera Type & Focal Length & Pixel Size & Resolution & DFOV &Spectral Range\\
\hline
Spectral camera & 16 mm & 5.5$\mu$m  & Spatial resolution: 2048 × 1088; spectral bands: 25 & 43.46° & 665–960 nm \\

Polarimetric camera & 16 mm & 3.45$\mu$m  & 2448 × 2048; polarization channels: 4 & 38.04° & 380–730 nm\\

Infrared camera & 19 mm & 12$\mu$m & 640 × 512  & 29° & 2.5–25$\mu$m \\
\hline
\end{tabular}
\end{table*}

\subsection{Multimodal Detection}
\begin{figure}[!t]
\centering
\includegraphics[width=\linewidth]{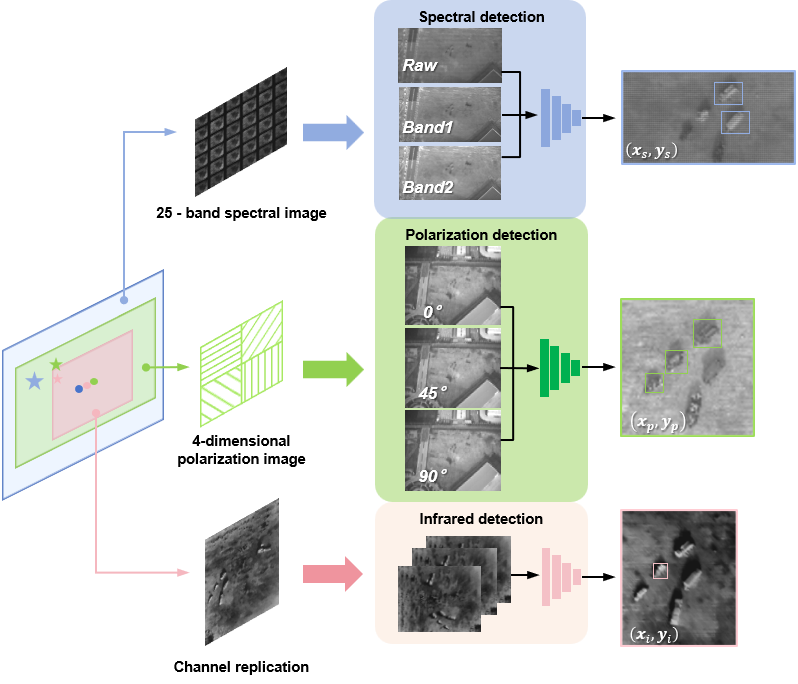}
\caption{Encode-then-detect paradigm for multimodal optical feature representation. The raw grayscale spectral image is decoupled into 25 spectral-band sub-images, from which two representative bands are selected and combined with the raw image to form a three-channel spectral input. For polarization imagery acquired at four analyzer orientations, the decomposed components at $0^\circ$, $45^\circ$, and $90^\circ$ are used to construct a three-channel polarization representation. For the infrared branch, the single-channel image is replicated into three channels to maintain a consistent input format.}
\label{fig_4}
\end{figure}

As illustrated in Fig.~\ref{fig_4}, the proposed framework exploits spectral, polarization, and infrared observations acquired from an overlapping field of view to detect multi-type camouflaged vehicle targets. To mitigate inter-modal misalignment and reduce redundant information across modalities, we adopt an optical-encoding-driven detection strategy. Specifically, for any modality $m \in \mathcal{M}$, the raw observation at time $t$ is first mapped into a compact representation through a modality-specific encoding function, i.e., 
\begin{equation}
E_t^{m}=\phi_m(X_t^{m}), \quad m\in\mathcal{M}.
\end{equation}
The encoded representation is then fed into the corresponding modality-specific detector to produce the detection output:
\begin{equation}
Y_t^{m}=F_m(E_t^{m}), \quad m\in\mathcal{M}.
\end{equation} 

This design strengthens modality-tailored feature extraction while producing modality-specific detection outputs for subsequent fusion.

\textbf{Spectral Encoding.} Multispectral data often contain substantial redundancy across bands. We therefore apply the OCF-CTBS band selection algorithm \cite{wang2024towards} to identify discriminative bands under diverse remote-sensing scenarios. To improve the generality of selected bands, each candidate band is assigned a weight according to its occurrence frequency across scenes, and bands with higher universality are retained. For target and background regions, we compute the mean intensity of each retained band to estimate band-wise sensitivity. Based on the sensitivity differences, we select the two bands that yield the largest target–background contrast, and stack these two band images with a reference intensity (or composite) image to form a three-channel spectrally encoded input.
\begin{equation}
E_t^{sp}=\left[I_t^{\mathrm{ref}},\,B_t^{(b_1)},\,B_t^{(b_2)}\right].
\end{equation}

\textbf{Polarization Encoding.} For snapshot polarization imagery acquired at four analyzer orientations, we use the decomposed polarization components at $0^\circ$, $45^\circ$, and $90^\circ$ to construct a compact three-channel representation: 
\begin{equation}
E_t^{po}=\left[P_t^{0^\circ},\,P_t^{45^\circ},\,P_t^{90^\circ}\right].
\end{equation}
This reduces input dimensionality and redundancy (e.g., the $135^\circ$ component is often correlated with the others via polarization constraints) while preserving discriminative polarization cues. The resulting encoded image helps the detector capture material-dependent differences (e.g., between metallic targets and non-metallic decoys) under camouflage.

\textbf{Infrared Encoding.} Infrared images are originally single-channel. To maintain a unified input format and enable direct use of standard detection backbones, we replicate the infrared channel three times to form a three-channel input. 
\begin{equation}
E_t^{ir}=\left[I_t^{ir},\,I_t^{ir},\,I_t^{ir}\right].
\end{equation}
This lightweight adaptation avoids modifying network architectures while retaining the thermal contrast information required for infrared-based detection.

\subsection{Cross-modal bounding box alignment}
\begin{figure}[!t]
\centering
\includegraphics[width=\linewidth]{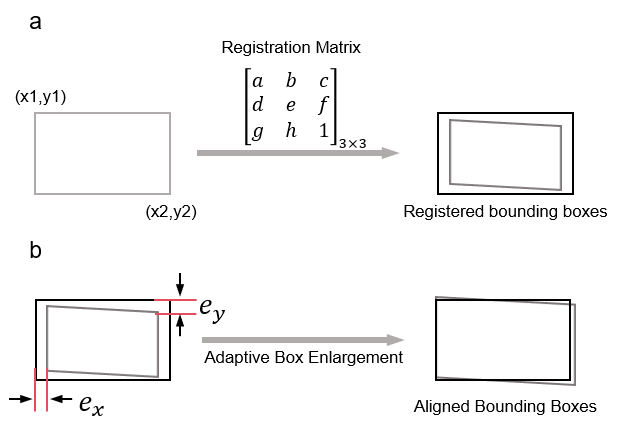}
\caption{Cross-modal bounding box alignment.}
\label{fig_5}
\end{figure}

Due to differences in spatial resolution, channel configuration, and FOV overlap across the multi-aperture sensors, establishing cross-modal spatial correspondences is necessary for reliable decision-level fusion. Rather than performing conventional pixel-/feature-level image registration, we propose a bounding-box-level adaptive registration strategy (Fig.~\ref{fig_5}), which is lightweight and suitable for real-time multi-modal detection.

Given three modality-specific detectors, we first obtain independent detection results (bounding boxes) for multispectral, polarization, and infrared images. We then perform a coarse cross-modal alignment in the box domain. Specifically, bounding boxes from non-reference modalities are projected into a unified reference coordinate system via homography transformation, as illustrated in Fig.~\ref{fig_5}(a).

\subsubsection{Homography-based box mapping}
As shown in Fig.~\ref{fig_5}(a), We select the polarization modality as the reference and map the remaining modalities to the polarization coordinate system using planar homographies estimated offline.

Let $\mathbf{H}$ be the $3 \times 3$ homography between the polarization image and another modality (e.g., multispectral), estimated from an offline calibration dataset. For a point $(u_1, v_1)$ in the polarization image, its corresponding location $(u_2, v_2)$ in the target modality is obtained by the standard homogeneous transformation, as defined in (\ref{eq:homography}) \cite{hartley2003multiple}.
\begin{equation}
\begin{bmatrix}
\tilde{u}_2 \\
\tilde{v}_2 \\
\tilde{w}_2
\end{bmatrix}
=
\mathbf{H}
\begin{bmatrix}
u_1 \\
v_1 \\
1
\end{bmatrix},
\qquad
u_2 = \frac{\tilde{u}_2}{\tilde{w}_2},
\qquad
v_2 = \frac{\tilde{v}_2}{\tilde{w}_2}.
\label{eq:homography}
\end{equation}

To map a detected bounding box, we transform its four corner points and then form an axis-aligned box in the reference coordinate system by taking the minimum and maximum mapped coordinates, as given in (\ref{eq:bbox_mapping}):
\begin{equation}
\begin{aligned}
x'_{\min} &= \min\nolimits_i x'_i, \qquad
y'_{\min} &= \min\nolimits_i y'_i,\\
x'_{\max} &= \max\nolimits_i x'_i, \qquad
y'_{\max} &= \max\nolimits_i y'_i.
\end{aligned}
\label{eq:bbox_mapping}
\end{equation}
where $(x'_1, y'_1)$ are the transformed corners. This box-level mapping dramatically reduces computation compared with image-level alignment, while providing sufficiently accurate correspondences for downstream fusion.

\subsubsection{Error sources and adaptive box expansion}

Although homography-based mapping is efficient, residual errors can shift mapped boxes, which is particularly detrimental for small targets whose mapped boxes may exhibit little or no overlap with their true locations. In our system, misalignment mainly comes from: (1) temporal asynchrony during multimodal acquisition that induces pixel displacement; and (2) imperfect feature correspondences during offline homography estimation.

To improve robustness, we introduce an adaptive bounding-box expansion strategy, as illustrated in Fig.~\ref{fig_5}(b). Assume the registration error is bounded by $e_x$ pixels horizontally and $e_y$ pixels vertically. Then the mapped location of any point $(x_0, y_0)$ may lie within:\[[x_0 - e_x, x_0 + e_x] \times [y_0 - e_y, y_0 + e_y],\]whose area is given by (\ref{eq:error_area}).
\begin{equation}
A_{\mathrm{err}} = (2e_x)(2e_y) = 4e_x e_y.
\label{eq:error_area}
\end{equation}
This error area provides a conservative criterion: when a detected box is too small, a plausible shift can eliminate cross-modal overlap. Let $A_b$ be the area of a mapped bounding box. If $A_b < A_{\mathrm{err}}$, we expand the box around its center to counteract misalignment; otherwise, no expansion is applied. A simple implementation is to enlarge the box by error margins as in (\ref{eq:margin_expand}) and then clip the resulting box to the image boundaries:
\begin{equation}
\begin{aligned}
x_{\min} &\leftarrow x_{\min} - e_x,\qquad
x_{\max} \leftarrow x_{\max} + e_x,\\
y_{\min} &\leftarrow y_{\min} - e_y,\qquad
y_{\max} \leftarrow y_{\max} + e_y.
\end{aligned}
\label{eq:margin_expand}
\end{equation}

Note that this module establishes correspondences for decision-level fusion and does not require pixel-wise dense registration.

\subsection{Multimodal Target Fusion via Consensus-Based Decision-Making}
\begin{figure*}[!t]
\centering
\includegraphics[width=\textwidth]{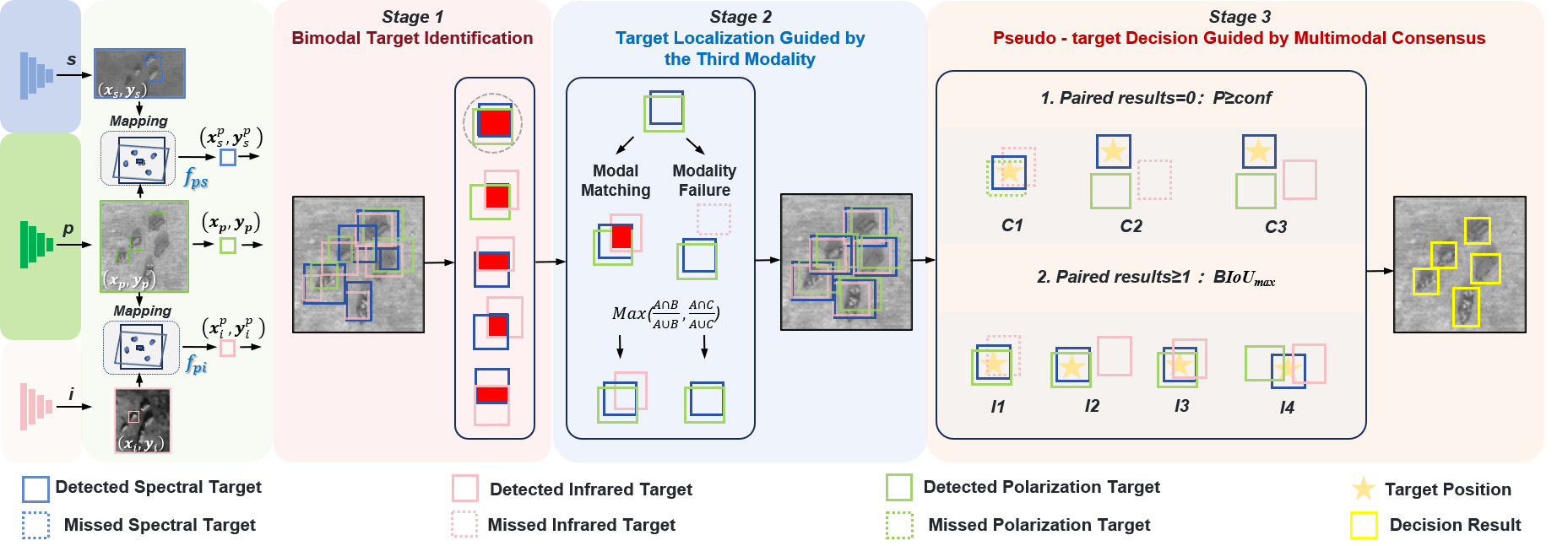}
\caption{Three-stage multimodal target fusion and decision strategy. In Stage 1, targets from two modalities are paired to identify the same object. In Stage 2, localization is guided by the third modality. In Stage 3, detection boxes are fused based on region-level consensus across modalities.}
\label{fig_6}
\end{figure*}

After mapping modality-specific detections into a unified coordinate system, a physical target may still yield multiple bounding boxes across modalities due to modality-dependent appearance and detector uncertainty. To robustly associate and consolidate these detections, we propose a three-stage fusion-and-decision pipeline, as illustrated in Fig.~\ref{fig_6} The pipeline (i) performs cross-modal box association for localization (Stage 1 and 2); and (ii) applies a consensus-guided decision rule (Stage 3) to suppress random false alarms and handle modality dropouts.

\subsubsection{Multimodal Fusion Localization}

The first two stages in Fig.~\ref{fig_6} are dedicated to target localization and multimodal bounding-box association.

Let $\mathcal{D}^{m}=\{(b_i^{m}, s_i^{m}, c_i^{m})\}$ denote detections from modality $m$, where
$b=[x_{\min}, y_{\min}, x_{\max}, y_{\max}]$, $s$ is the confidence score, and $c$ is the predicted category.
Cross-modal association is performed only between different modalities.

\textit{Stage 1: Bimodal pairing via cross-modal matching.}
We compute IoU between boxes of the same category from two modalities $m_1$ and $m_2$ to form an IoU matrix $\mathbf{M}^{m_1 m_2}$, e.g., $\mathbf{M}^{PI}$, $\mathbf{M}^{IS}$, and $\mathbf{M}^{SP}$.
Pairing is then obtained by solving a one-to-one matching problem under an IoU criterion, i.e., box $b_i^{m_1}$ and box $b_j^{m_2}$ are matched only if their IoU is maximal among feasible candidates.

In practice, we use a greedy maximum-IoU matching with suppression.
Specifically, we repeatedly select the candidate pair with the maximum IoU between two modalities, add it to the matched set, and remove the paired boxes from subsequent consideration.
Alternatively, a Hungarian assignment with cost $1-\mathrm{IoU}$ can also be used for global optimization.
This bimodal pairing strategy mitigates mismatches for spatially adjacent targets by enforcing one-to-one correspondences.

\textit{Stage 2: Completion to trimodal association.}
In the second stage, for each bimodal pair $b_i^{m_1}$ and $b_j^{m_2}$, we form a provisional support region $b_{ij}^{m_1 m_2}$ (e.g., the union box or a weighted fusion of the two boxes).
We then match boxes from the third modality $m_3$ according to (\ref{eq:trimodal_match}).

\begin{equation}
j^{*}=\arg\max_j \mathrm{IoU}\!\left(b_j^{m_3},\, b_{ij}^{m_1 m_2}\right).
\label{eq:trimodal_match}
\end{equation}

If no box satisfies $\mathrm{IoU}>0$, the pair remains bimodal rather than being discarded, which improves robustness to modality degradation.
We then add the selected box, if any, to the matched set and remove the associated boxes from further consideration in subsequent matching.

\subsubsection{Multimodal Consensus-Guided Camouflage Target Decision}
In the third stage, to fully exploit complementary multimodal detection results, we propose a multimodal consensus-guided camouflage target decision strategy.

Each modality-specific detection model exhibits a certain degree of generalization capability across target categories, while being sensitive to different physical attributes. For example, the infrared detector is particularly responsive to thermal radiation patterns. Consequently, detection results from different modalities are complementary and largely independent, and each model has a relatively low probability of producing identical false detections for the same target. To mitigate random false positives introduced by any single modality, we introduce a multimodal regional consensus voting mechanism, in which each modality contributes equally to the final decision. As shown in Stage 3 of Fig.~\ref{fig_6}, the number of bounding boxes within a region reflects the voting outcome of the three detection models. Two cases are considered:

For each associated group $g=\{S,P,I\}$, let $V_g$ denote the number of modalities that produce a detection within the same support region (i.e., $V_g \in \{1,2,3\}$). We adopt the following decision rule:

\begin{itemize}
\item \textit{No cross-modal agreement} $(V_g=1)$: the group is rejected by default. To avoid suppressing highly reliable single-modality detections, we retain the group if its confidence exceeds a high threshold $\theta_H$, i.e., $s^m > \theta_H$, in which case it contributes a single vote.

\item \textit{Cross-modal agreement} $(V_g \geq 2)$: the group is accepted as a valid target, and the corresponding boxes are fused to produce a single output according to (\ref{eq:fuse_rule}):

\begin{equation}
D_g = \mathcal{F}\!\left(\{b^m, s^m\}_{m \in g}\right).
\label{eq:fuse_rule}
\end{equation}
\end{itemize}

By leveraging multimodal consensus rather than relying on any single detector, the proposed strategy effectively reduces false alarms and improves camouflage target recognition in complex environments.

\textit{Weighted box fusion:}
Given boxes $\{b^m\}_{m \in g}$ with confidences $\{s^m\}_{m \in g}$, we fuse them via confidence-weighted aggregation, where the modality weight is defined in (\ref{eq:weight_def}):
\begin{equation}
w^m =
\begin{cases}
\phi(s^m), & \text{if modality } m \text{ is available}\\
0, & \text{otherwise}
\end{cases}
\label{eq:weight_def}
\end{equation}

\begin{equation}
b_g =
\frac{\sum\limits_{m \in \{S,P,I\}} w^m b^m}
{\sum\limits_{m \in \{S,P,I\}} w^m},
\label{eq:box_fusion}
\end{equation}

\begin{equation}
s_g =
\frac{\sum\limits_{m \in \{S,P,I\}} w^m s^m}
{\sum\limits_{m \in \{S,P,I\}} w^m},
\label{eq:score_fusion}
\end{equation}
where $\phi(\cdot)$ can be the identity function or a calibrated mapping (e.g., $\phi(s)=s$), and the fused box and score are computed by (\ref{eq:box_fusion}) and (\ref{eq:score_fusion}), respectively. This fusion removes redundant detections while leveraging complementary information across modalities.

The pseudocode for Stages 1--3 is shown in Algorithm~1.

\begin{algorithm}[t]
\caption{Consensus-Guided Multimodal Target Fusion}
\label{alg:consensus_fusion}
\begin{algorithmic}[1]
\STATE \textbf{Input:} Aligned detection sets $\{\mathcal{D}_k^m\}_{m\in\{S,P,I\}}$ for category $k$; threshold $\theta_H$
\STATE \textbf{Output:} Fused detection set $\mathcal{F}_k$
\STATE Initialize $\mathcal{F}_k \leftarrow \emptyset$
\STATE Generate associated groups $\mathcal{G}$ by bimodal pairing and trimodal completion according to Stages 1--2
\FOR{each group $g \in \mathcal{G}$}
    \STATE Let $\mathcal{M}_g$ be the contributing modality set and $V_g = |\mathcal{M}_g|$
    \IF{$V_g = 1$}
        \STATE Let the only detection in $g$ be $(b^m, s^m, k)$
        \IF{$s^m > \theta_H$}
            \STATE $\mathcal{F}_k \leftarrow \mathcal{F}_k \cup \{(b^m, s^m, k)\}$
        \ENDIF
    \ELSE
        \STATE Compute modality weights $w^m$ according to (\ref{eq:weight_def})
        \STATE Compute fused box $b_g$ and score $s_g$ according to (\ref{eq:box_fusion}) and (\ref{eq:score_fusion})
        \STATE $\mathcal{F}_k \leftarrow \mathcal{F}_k \cup \{(b_g, s_g, k)\}$
    \ENDIF
\ENDFOR
\STATE \textbf{return} $\mathcal{F}_k$
\end{algorithmic}
\end{algorithm}

\section{Experiments}
\subsection{Multimodal CVD Dataset}
\begin{figure}[!t]
\centering
\includegraphics[width=\linewidth]{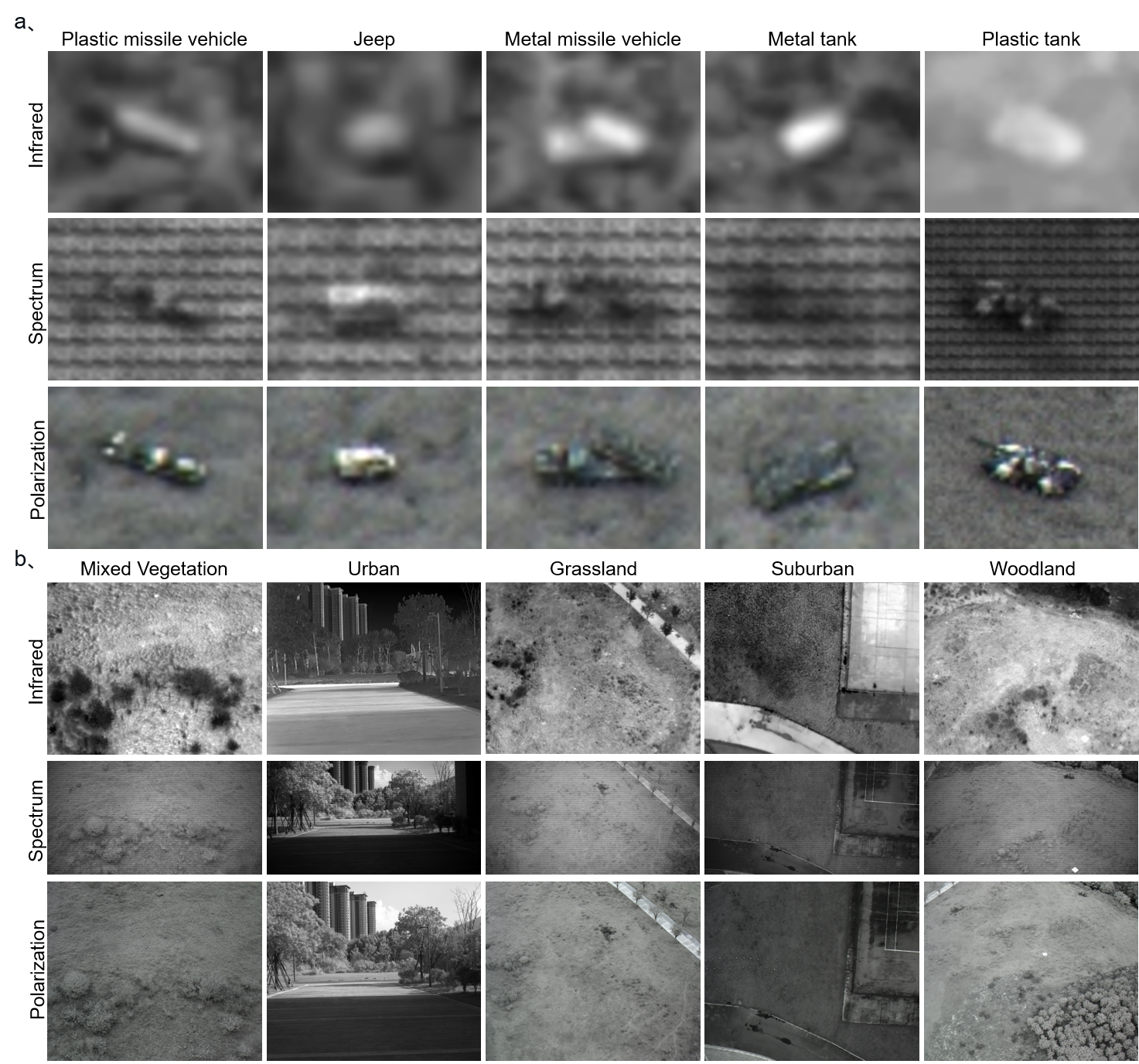}
\caption{Multimodal camouflaged target image dataset. (a) Enlarged local crops of representative spectral, polarimetric, and infrared images of camouflaged targets. (b) Scene images showing the environments in which the targets are located, including vegetated areas, urban road/building surroundings, and open bare-ground scenes, where the targets usually appear as small and weakly distinguishable objects under complex backgrounds.}
\label{fig_7}
\end{figure}

This subsection introduces the multimodal camouflaged vehicle detection (CVD) dataset used for experimental evaluation. Fig.~\ref{fig_7} presents representative examples from the proposed multimodal CVD dataset and the complex scenarios it covers. The dataset contains synchronized observations from three modalities, including infrared, polarization, and multispectral imaging. All modalities are acquired using a unified capture procedure with temporal synchronization and consistent viewing geometry, so that samples are indexed in a one-to-one manner across modalities. The dataset contains a total of 3,726 multimodal image samples covering five camouflaged vehicle categories, and is divided into training and test sets with a ratio of 4:1.

The target patches shown in Fig.~\ref{fig_7}(a) are enlarged local crops for visual clarity, rather than full images at the original scale. As summarized in Table~\ref{table_2}, the targets occupy only 0.0139\%--0.0305\% of the image area in polarimetric images.

\begin{table*}[!t]
\renewcommand{\arraystretch}{1.2}
\caption{Target Categories and Image Statistics of the Multimodal Camouflaged Vehicle Dataset}
\label{table_2}
\centering
\begin{tabular}{c c c c c c}
\hline
Target Type & Metal Tank & Plastic Tank & Metal Missile Vehicle & Plastic Missile Vehicle & Jeep\\
\hline
Number of images (three modalities) & 2994 & 2924 & 2063 & 1834 & 1439 \\
Target area ratio in polarimetric images (\%) & 0.0237 & 0.0195 & 0.0305 & 0.0263 & 0.0139\\
\hline
\end{tabular}
\end{table*}

Due to modality-dependent imaging mechanisms, the target boundary appearance can vary across modalities. Therefore, we annotate target instances independently for each modality to better reflect modality-specific characteristics while preserving cross-modal sample correspondence. The dataset is split into training and test sets to facilitate evaluation under complex backgrounds and cross-modal conditions.

The dataset includes five types of camouflaged vehicle targets: metal tank, plastic tank, metal missile vehicle, plastic missile vehicle, and jeep. These categories exhibit clear differences in material properties, geometric scale, and surface reflectance, which lead to distinct responses across spectral, polarization, and infrared modalities. Fig.~\ref{fig_7}(a) further illustrates the corresponding non-visible observations. Infrared imaging highlights radiation contrast induced by material-dependent thermal behavior and temperature differences; polarization imaging reflects differences in surface reflection properties that can enhance structural edges; and multispectral imaging captures material-dependent spectral signatures. These cues are complementary and jointly facilitate robust camouflage recognition. 

Fig.~\ref{fig_7}(b) shows typical scenes covered by the dataset, including mixed vegetation, urban environments, grassland, suburban areas, and woodland. The scenes differ significantly in background texture density, illumination conditions, and environmental complexity, enabling a comprehensive evaluation of detection robustness in realistic complex environments.The dataset provides overlapping fields of view and temporal synchronization across modalities without requiring pixel-wise perfect registration, reflecting practical multimodal acquisition conditions.

\subsection{Evaluation Metrics}
To comprehensively assess detection performance, we adopt mean Average Precision (mAP), Precision (P), and Recall (R) as evaluation metrics. Among them, mAP@0.5 and mAP@0.5:0.95 are widely used in object detection benchmarks and measure performance under different localization accuracy requirements. Specifically, P and R evaluate the accuracy and completeness of the predictions, respectively. AP summarizes the P–R relationship for a single class by integrating the P–R curve. For multi-class detection, mAP is obtained by averaging AP over all classes.

\subsection{Implementation Details}
All detection models are implemented based on the YOLOv5 architecture \cite{redmon2016you}. Model training, validation, and offline accuracy evaluation are performed on an NVIDIA RTX 3090 GPU. For deployment, the proposed method is executed on an onboard NVIDIA Jetson AGX Orin platform, on which the end-to-end inference latency is 16 ms. For the polarization and spectral detection networks, we adopt the Adam optimizer with the momentum parameter set to 0.99, while the infrared detection network is optimized using stochastic gradient descent (SGD). The initial learning rate is set to 0.01 and is linearly decayed to 0.0001 over the training process. The batch size is fixed to 16, and all models are trained for 300 epochs. The model checkpoint achieving the best performance on the validation set is selected for inference.

During testing, the confidence threshold is set to 0.1, while the confidence threshold for retaining single-modality detections in the proposed multimodal decision module is set to 0.75.

\subsection{Camouflaged Vehicle Detection}
To validate the effectiveness of the proposed approach, we conduct comparisons on our presented multimodal camouflaged vehicle dataset against (1) unimodal baselines and (2) representative feature-level bimodal fusion (mid-level fusion) methods. For unimodal detection, we employ YOLOv5s trained and tested on each individual modality. In addition, we evaluate an early-fusion baseline by stacking all modality images into a multi-channel input tensor and modifying only the input convolution to match the channel number. Due to the limited availability of open-source multimodal detection frameworks, we select several recent and representative multimodal methods for comparison, including CTF \cite{qingyun2021cross}, ICAF \cite{shen2024icafusion}, and DEYOLO \cite{chen2024deyolo}.

\begin{figure*}[!t]
\centering
\includegraphics[width=\linewidth]{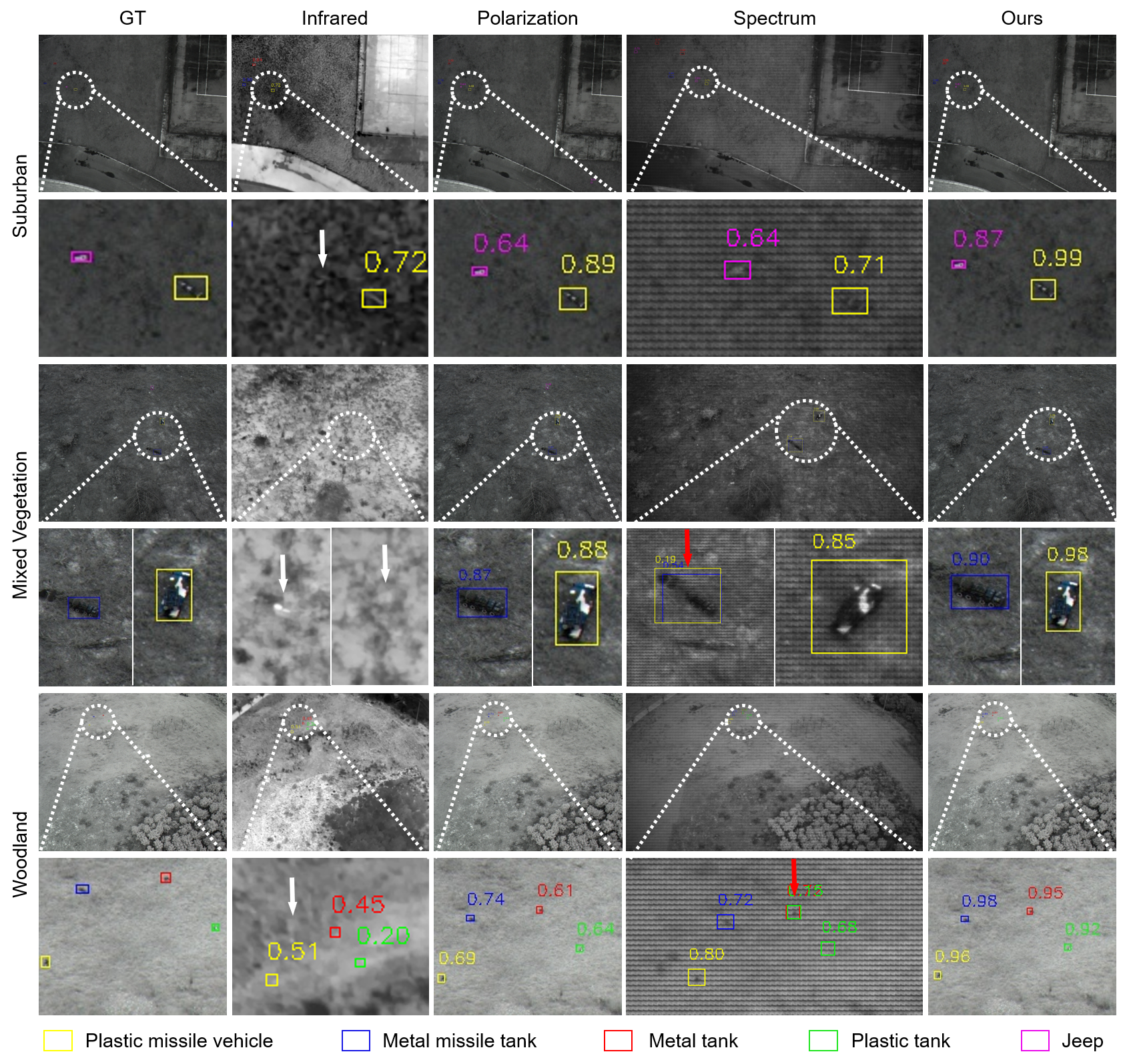}
\caption{Qualitative comparison of our method and single-modality methods across scenes. Colored boxes denote object categories. White dashed circles denote zoomed-in regions. White arrows indicate missed targets, while red arrows denote false alarms, highlighting typical unimodal failures and the improved robustness of our method.}
\label{fig_8}
\end{figure*}

As illustrated in Fig.~\ref{fig_8}, each modality exhibits inherent limitations under different scenarios. For example, in the Metal tank case, the infrared modality fails to detect the target. This is mainly because infrared imaging depends on the surface-emitted radiation, which is influenced by temperature and emissivity. Metals typically have low emissivity (approximately 0.05–0.3) and reflect a substantial portion of incident infrared radiation, which can reduce the apparent thermal contrast and lead to missed detections.

In the Mixed Vegetation scenario, the spectral modality is affected by complex vegetation textures and material similarity, causing false detections of the Plastic missile vehicle. Similarly, in the Woodland scenario, the spectral modality produces noticeable false detections of the Plastic tank, suggesting that spectral cues alone can be disturbed by natural texture patterns and reflectance variations in complex backgrounds. Overall, unimodal YOLOv5 detectors show different sensitivities across target types and environments: although unimodal models can sometimes improve recall by being highly responsive to certain targets, they are also more prone to false alarms when modality-specific cues are unreliable.

\begin{figure*}[!t]
\centering
\includegraphics[width=\linewidth]{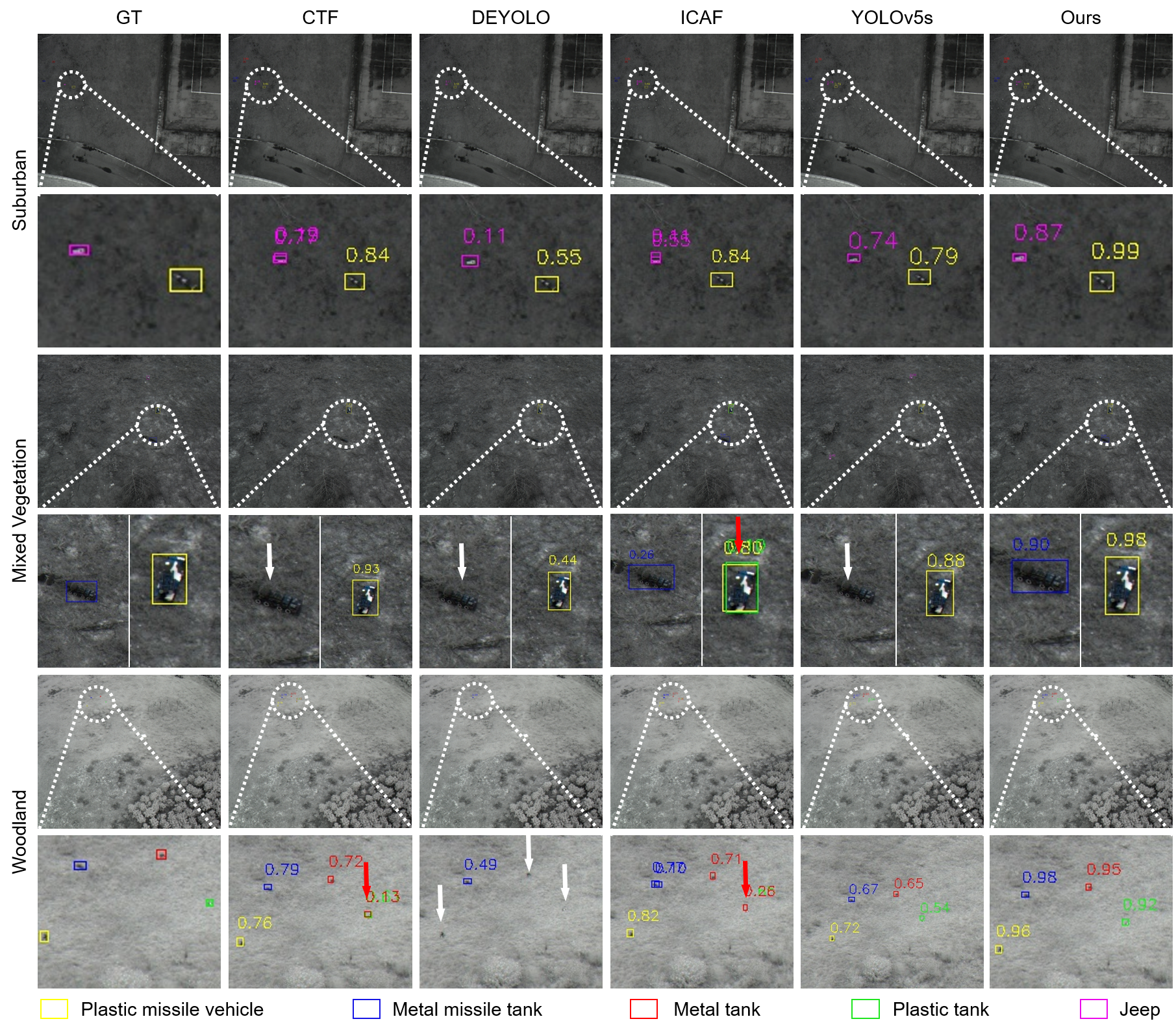}
\caption{Qualitative comparison between our method and other multi-modality methods across different scenes. Colored boxes denote object categories. White dashed circles denote zoomed-in regions. White arrows indicate missed targets, while red arrows denote false alarms, highlighting typical unimodal failures and the improved robustness of our method.}
\label{fig_9}
\end{figure*}

In contrast, as shown in Fig.~\ref{fig_9}, when multimodal images are stacked and fed into YOLOv5 (early fusion), only the input convolution kernel channels are changed, resulting in a minor parameter increase (about 0.26\%). This simple strategy yields a clear gain in detection accuracy: compared with the best unimodal baseline, mAP@0.5:0.95 improves by 11.2\%, demonstrating the complementarity of multimodal information and its benefits for robust detection.

Besides, the feature-level fusion models such as CTF, DEYOLO, and ICAF still suffer from significant missed detections and false alarms in complex scenes (e.g., Woodland and Mixed Vegetation). This is primarily because multimodal feature fusion can be affected by cross-modality inconsistency and information redundancy/exclusivity, limiting the ability to fully exploit complementary cues. DEYOLO exhibits more missed detections, largely because pixel-level misalignment between modalities makes it difficult to establish reliable cross-modal correspondences when relying mainly on convolutional receptive fields, which can cause target features to be weakened or lost. CTF improves detection by leveraging transformers to establish global cross-modal interactions, but it does not completely resolve modality redundancy and exclusivity, leading to residual false detections. These results indicate that performance gains from feature-level fusion are highly dependent on fusion architecture design and may remain limited under challenging alignment and background conditions.

By contrast, the proposed approach introduces a trimodal decision-level fusion strategy. The collaborative trimodal decision process suppresses false alarms and missed detections caused by unimodal weaknesses, producing higher and more stable prediction confidence across scenarios. In particular, under strong camouflage and complex backgrounds, decision-level fusion provides more comprehensive decision criteria, enabling the detector to maintain high recall while effectively reducing false detections, thereby improving stability and reliability.

\begin{table}[!t]
\renewcommand{\arraystretch}{1.1}
\setlength{\tabcolsep}{2pt}
\caption{Performance comparison of ours with single-modality and multimodal methods.}
\label{table_3}
\centering
\begin{tabular}{lccccc}
\hline
Method & Modality & mAP$_{0.5}$ & mAP$_{0.5:0.95}$ & P & R \\
\hline
YOLOv5s-PZ                  & 1 & 0.772 & 0.376 & 0.917 & 0.616 \\
YOLOv5s-IR                  & 1 & 0.731 & 0.336 & 0.873 & 0.522 \\
YOLOv5s-SP                  & 1 & 0.757 & 0.372 & 0.923 & 0.586 \\
YOLOv5s                     & 3 & 0.783 & 0.418 & 0.921 & 0.636 \\
CTF~\cite{qingyun2021cross} & 3 & 0.769 & 0.361 & 0.840 & 0.683 \\
ICAF~\cite{shen2024icafusion} & 3 & 0.771 & 0.356 & 0.881 & 0.654 \\
DEYOLO~\cite{chen2024deyolo}  & 3 & 0.734 & 0.433 & 0.969 & 0.487 \\
\textbf{Ours}   & \textbf{3} & \textbf{0.827} & \textbf{0.437} & \textbf{0.899} & \textbf{0.704} \\
\hline
\end{tabular}
\end{table}

Quantitatively, experiments on five classes of camouflaged vehicle targets (Table~\ref{table_3}) show that our method achieves the best overall detection performance. Compared with unimodal baselines, our method improves mAP by 5.5\%, and it outperforms the best multimodal baseline by 5.6\%. The results also suggest that feature-level fusion is more sensitive to pixel misalignment, which can cause inconsistent feature extraction across modalities. Moreover, most prior multimodal methods focus on visible–infrared bimodal fusion and do not fully consider either (1) richer inter-modality interactions in trimodal settings or (2) preserving intra-modality specificity for modalities such as spectral, polarization, and infrared. In this context, decision-level fusion provides a practical and extensible solution for arbitrary modality combinations, making it particularly suitable for real-world multimodal detection tasks.

\subsection{Analysis on the Multimodal Fusion Decision Strategy}
Since the NMS module in YOLOv5 \cite{redmon2016you} also performs box filtering, we compare the proposed fusion decision strategy against NMS-based fusion to quantify its effectiveness. Results are summarized in Table~\ref{table_4}. With a small NMS IoU threshold, only the highest-confidence box is retained in each region, yielding high precision but low recall. Increasing the IoU threshold preserves more modality-specific boxes, substantially improving recall. Precision may increase slightly at first (by recovering valid boxes), but then degrades as redundant and low-quality boxes accumulate. Overall, using NMS alone to fuse multimodal outputs leads to clear performance degradation across metrics. Even at its best IoU threshold, the resulting mAP is over 40\% lower than that achieved by the proposed fusion decision strategy.

\begin{table}[!t]
\renewcommand{\arraystretch}{1.1}
\setlength{\tabcolsep}{3pt}
\caption{Performance comparison between the proposed consensus decision fusion and standard NMS under different IoU threshold settings}
\label{table_4}
\centering
\small
\resizebox{\columnwidth}{!}{%
\begin{tabular}{c c c c c }
\hline
Method & P & R & mAP$_{0.5}$ & mAP$_{0.5:0.95}$\\
\hline
Standard NMS ($\tau_{\mathrm{nms}}=10^{-8}$) & 0.436 & 0.309 & 0.301 & 0.165 \\
Standard NMS ($\tau_{\mathrm{nms}}=0.9$) & 0.402 & 0.708 & 0.388 & 0.204 \\
Standard NMS (optimal, $\tau_{\mathrm{nms}}=0.4$) & 0.460 & 0.694 & 0.433 & 0.222 \\
\textbf{Consensus decision fusion (ours)} & \textbf{0.899} & \textbf{0.704} & \textbf{0.827} & \textbf{0.437} \\
\hline
\end{tabular}%
}
\end{table}

This gap stems from three main causes. First, NMS effectively selects boxes from a single modality in a competitive manner, which can amplify modality-specific false positives in cluttered scenes. In contrast, our consensus-based fusion suppresses false alarms by requiring cross-modality agreement. Second, NMS applies a uniform suppression rule and is therefore sensitive to inter-modal misregistration; small alignment errors can trigger incorrect suppression or retention. Third, NMS does not explicitly regulate how many boxes are discarded within the same target region. Our method instead enforces structured cross-modal pairing, retaining only the most compatible detections. This prevents erroneous suppression that can induce localization drift or missed detections, thereby improving robustness in complex multimodal settings.

\subsection{Ablation Study}

To quantitatively evaluate the effectiveness of the proposed adaptive registration scaling strategy and to verify the necessity of multimodal integration within the overall framework, a series of ablation studies were conducted.

First, we compare the complete three-modality framework with a single-modality baseline (Polarization only) while keeping the network architecture unchanged. As reported in Table~\ref{table_5}, the full multimodal configuration achieves an improvement of 0.089 in mAP@0.5 and 0.098 in mAP@0.5:0.95 over the single-modality counterpart. These results demonstrate that complementary spectral and infrared cues provide additional discriminative information beyond polarization, which is essential for robust detection of camouflaged vehicles in complex environments.

\begin{table}[!t]
\renewcommand{\arraystretch}{1.2}
\caption{Polarization-Only vs Full Spectral–Polarization–Infrared Framework.}
\label{table_5}
\centering
\begin{tabular}{c c c c c}
\hline
Method & mAP$_{0.5}$ & mAP$_{0.5:0.95}$ & P & R \\
\hline
Polarization only & 0.738 & 0.339 & 0.877 & 0.577 \\
\textbf{Ours} & \textbf{0.827} & \textbf{0.437} & \textbf{0.899} & \textbf{0.704} \\
\hline
\end{tabular}
\end{table}

Second, we evaluate the contribution of the adaptive scaling mechanism in the registration module. Specifically, we remove the proposed scaling strategy while keeping all other components identical. As shown in Table~\ref{table_6}, introducing the scaling operation leads to improvements of 0.567 in mAP@0.5 and 0.338 in mAP@0.5:0.95. This substantial performance gain verifies that the adaptive scaling strategy effectively compensates for spatial misalignment and scale inconsistency across modalities, thereby improving localization accuracy and overall detection performance.

\begin{table}[!t]
\renewcommand{\arraystretch}{1.2}
\caption{Ablation study on the adaptive registration scaling strategy.}
\label{table_6}
\centering
\begin{tabular}{c c c c c}
\hline
Method & mAP$_{0.5}$ & mAP$_{0.5:0.95}$ & P & R \\
\hline
Unscaled & 0.260 & 0.099 & 0.376 & 0.393 \\
\textbf{Ours} & \textbf{0.827} & \textbf{0.437} & \textbf{0.899} & \textbf{0.704} \\
\hline
\end{tabular}
\end{table}

Since the modality-specific detectors are replaceable backbone components, we focus here on the additional computational overhead introduced by the proposed post-detection modules. As reported in Table~\ref{table_7}, box-level registration and consensus decision fusion introduce no learnable parameters and only limited latency overhead, with average processing times of 0.1996 ms and 0.0477 ms per synchronized multimodal sample, respectively. These results indicate that the proposed post-detection design remains lightweight and can be integrated with different detector backbones in a plug-and-play manner.

\begin{table}[!t]
\renewcommand{\arraystretch}{1.2}
\caption{Computational overhead of the proposed post-detection modules.}
\label{table_7}
\centering
\begin{tabular}{lcc}
\hline
Module & Learnable Params & Latency (ms) \\
\hline
Box-level registration & 0 & 0.1996 \\
Consensus decision fusion & 0 & 0.0477 \\
\hline
\end{tabular}
\end{table}

\subsection{Detection on an Additional Multimodal Dataset}
To assess generalization to arbitrary modality combinations, we further evaluate on the public MID-1K dataset, which provides RGB–Depth–Infrared modalities \cite{mota2024multimodal}. MID-1K contains diverse indoor scenes with pedestrians as the primary targets. Importantly, our method is applied without any dataset-specific architectural changes or modality-dependent customization, i.e., the same unified detection and decision-level fusion framework is used as in previous experiments.

\begin{figure}[!t]
\centering
\includegraphics[width=\linewidth]{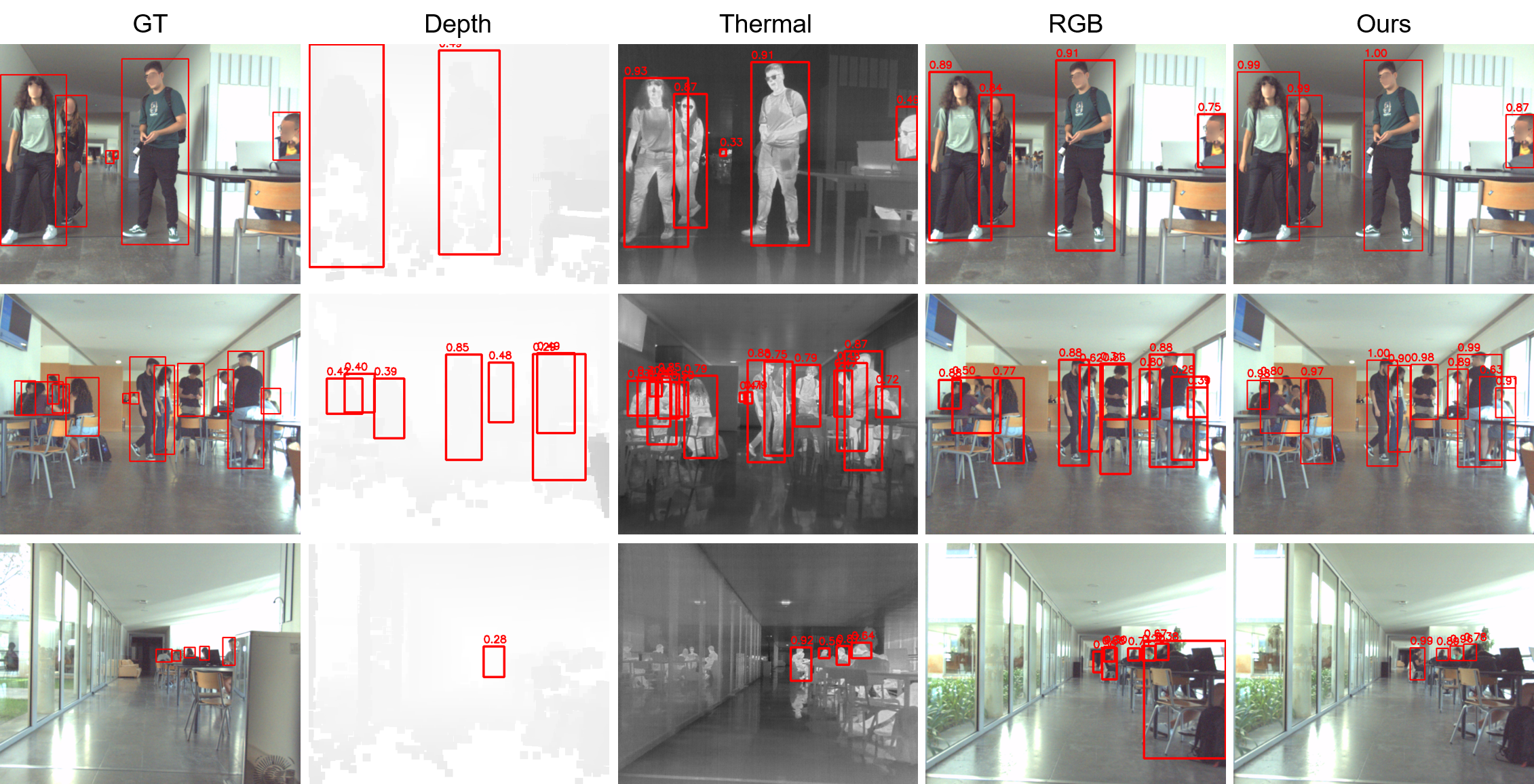}
\caption{Multimodal detection results on the MID-1K dataset.}
\label{fig_10}
\end{figure}

Qualitative results in Fig.~\ref{fig_10} show that MID-1K poses substantial challenges, including dense crowds, large scale variation, and strong inter-modal discrepancies. In particular, depth maps frequently exhibit blurred contours, while RGB images are sensitive to illumination non-uniformity, leading to unstable unimodal predictions. In contrast, our method produces more complete detections with consistent target counts and improved robustness under occlusion, demonstrating clear benefits from cross-modal complementarity.

Table~\ref{tab_mid1k_compare} compares our method with published results on MID-1K. The published work also adopts YOLOv5s as the baseline detector, making its detector setting broadly comparable to ours. Their results show that, among single modalities, thermal achieves the best performance, while DRT provides the strongest early-fusion baseline with an mAP$_{0.5}$ of 0.531 and an mAP$_{0.5:0.95}$ of 0.590. A decision-level late-fusion scheme based on MLP further improves the stricter metric to 0.631 mAP$_{0.5:0.95}$, showing the potential of post-detection fusion on MID-1K. Under the same three-modality setting, our method achieves a higher mAP$_{0.5}$ of 0.588, outperforming the published DRT baseline by 5.7\%. These results verify that the proposed framework remains effective on an additional public multimodal dataset and can generalize well across different sensing conditions.

\begin{table}[!t]
\renewcommand{\arraystretch}{1.08}
\setlength{\tabcolsep}{2.5pt}
\caption{Comparison with published results on the MID-1K dataset.}
\label{tab_mid1k_compare}
\centering
\begin{tabular}{lcccc}
\hline
Method & Fusion level & Modality & mAP$_{0.5}$ & mAP$_{0.5:0.95}$ \\
\hline
Depth                & Single         & 1 & 0.209 & 0.248 \\
RGB                  & Single         & 1 & 0.291 & 0.330 \\
Thermal              & Single         & 1 & 0.412 & 0.482 \\
DRT                  & Early fusion   & 3 & 0.531 & 0.590 \\
Late Fusion with MLP & Late fusion    & 3 & 0.381 & 0.631 \\
\textbf{Ours}        & \textbf{Decision fusion} & 3 & \textbf{0.588} & \textbf{0.290} \\
\hline
\end{tabular}
\end{table}

The performance under the stricter mAP@0.5:0.95 metric drops noticeably. This suggests that while the proposed method improves target discoverability and coarse localization, achieving highly accurate box regression under stricter IoU thresholds remains challenging on MID-1K. This is likely related to the dataset’s uneven scale distribution and the high proportion of small instances, where minor localization deviations are penalized more heavily. Nevertheless, the consistent advantage at mAP@0.5 verifies the robustness and scalability of the proposed framework across different sensing configurations and arbitrary modality combinations, supporting its practicality for real-world multimodal detection.

\section{Conclusion}
This work addresses camouflaged target detection in complex scenes and focuses on two practical bottlenecks of multimodal systems: the lack of accurate inter-modal registration and the trade-off between detection accuracy and real-time efficiency \cite{xu2023murf}. Unlike conventional pipelines based on pixel-level registration and feature-level fusion on aligned inputs, we co-design the imaging hardware and detection algorithm and propose a real-time multimodal camouflaged-target detection framework based on a multi-aperture imaging system. The proposed framework offers three main advantages: it avoids precise pixel alignment through a consensus-based decision module that can be readily integrated with arbitrary detectors, it exploits spectral and polarization cues beyond shape priors to characterize camouflage materials and surface textures \cite{he2023camouflaged,lyu2023shape}, and it enables robust collaborative decision-making through distinctive and complementary modality-specific outputs without manually tuned modality weights.

Despite these advantages, several limitations remain. Although the constructed multimodal dataset includes multiple target categories, varying scales, and data acquired by different multi-aperture imaging systems, the gap between scaled models and real deployments cannot be fully eliminated. In addition, performance on higher-speed platforms may be affected by motion blur, severe illumination changes, and cloud or fog occlusion. Nevertheless, the proposed framework indicates encouraging generalization capability. Each branch operates on modality-discriminative inputs, such as selected spectral bands and encoded polarization features, which helps reduce the ambiguity of unimodal observations. Moreover, experiments on MID-1K demonstrate that the method can be transferred to a different application domain without architectural modification \cite{mota2024multimodal}. Future work will extend the framework to real camouflaged targets, include more challenging scenes and platform conditions, and further improve robustness through platform-specific calibration and lightweight fine-tuning.

\section*{Acknowledgments}
The authors would like to acknowledge National Natural Science Foundation of China (62575072)

\bibliographystyle{IEEEtran}
\bibliography{refs}

\end{document}